\documentclass[conference]{IEEEtran}  

\IEEEoverridecommandlockouts

\usepackage{cite}
\usepackage{amsmath,amssymb,amsfonts}
\usepackage{algorithmic}
\usepackage{graphicx}
\usepackage{textcomp}
\usepackage{xcolor}
\def\BibTeX{{\rm B\kern-.05em{\sc i\kern-.025em b}\kern-.08em
    T\kern-.1667em\lower.7ex\hbox{E}\kern-.125emX}}
    
\usepackage{caption, subcaption}

\def\bX{\mathbf{X}}
\def\bx{\mathbf{x}}
\def\bmu{\boldsymbol{\mu}}
\def\bw{\mathbf{w}}

\begin{document}

\title{Comparison of Classification Algorithms Towards Subject-Specific and Subject-Independent BCI 
}

\author{\IEEEauthorblockN {Parisa Ghane, Narges Zarnaghi Naghsh, and Ulisses Braga-Neto}
\IEEEauthorblockA{Dept.\ of Electrical and Computer Engineering \\
Texas A\&M University\\
College Station, United States \\
ulisses@tamu.edu}

\thanks{\emph {Preprint.}}
}


\maketitle

\begin{abstract}
Motor imagery brain computer interface designs are considered difficult due to limitations in subject-specific data collection and calibration, as well as demanding system adaptation requirements. 
Recently, subject-independent (SI) designs received attention because of their possible applicability to multiple users without prior calibration and rigorous system adaptation. SI designs are challenging and have shown low accuracy in the literature. 
Two major factors in system performance are the classification algorithm and the quality of available data. This paper presents a comparative study of classification performance for both SS and SI paradigms. 
Our results show that classification algorithms for SS models display large variance in performance. Therefore, distinct classification algorithms per subject may be required. SI models display lower variance in performance but should only be used if a relatively large sample size is available. 
For SI models, LDA and CART had the highest accuracy for small and moderate sample size, respectively, whereas we hypothesize that SVM would be superior to the other classifiers if large  training sample-size was available. 
Additionally, one should choose the design approach considering the users. While the SS design sound more promising for a specific subject, an SI approach can be more convenient for mentally or physically challenged users. 
\end{abstract}


\section{INTRODUCTION}
Electroencephalography (EEG) pattern recognition is an essential component of Brain Computer Interface (BCI) systems.
In BCI systems, the patterns in recorded EEG signals are identified, classified, and finally translated to a control signal to be sent to an external device.
In particular, motor imagery (MI) EEG-based BCI systems are believed to show the brain activity in regions that are involved in movement imagination, without real limb execution \cite{pfurtscheller2001motor, jeannerod1995mental, decety1994mapping}.
This application is beneficial to the neuro-rehabilitation of a large community of paralyzed patients whose corticospinal track is blocked.

\begin{figure}
    \centering
  \includegraphics[scale=0.45]{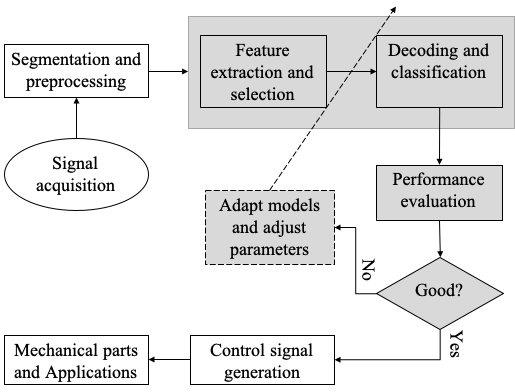}
    \caption{Basic parts of an offline BCI system. The pattern recognition parts are shaded.}
    \label{fig:bci}
\end{figure}

The EEG temporal and spatial characteristics are correlated with an individual's physical and mental state and may vary between subjects. Hence, subject-specific (SS) BCI models are designed and trained per subject, which makes SS BCI designs time consuming and inconvenient.
To overcome the limitations of BCI, subject-independent (SI) BCI designs were introduced and received attention recently. The purpose of SI models is to train a general model that can be used by new subjects with little or no experimental calibration and model parameter adaptation. 

Although MI EEG-based BCI is a major research area, it is not yet fully understood, possibly due to the lack of reliable data as well as complicated data collection and model training. Studies on SS BCI designs show that EEG classification on MI data has considerably lower accuracy than some other BCI tasks, such as P300, ERP, or VEP \cite{kevric2017comparison, guger2009many}. For example, a recent study \cite{lee2019eeg} reported a mean classification accuracy of $71.1\%$ for MI data, but $96.7\%$ and $95.1\%$ for the ERP and SSVEP paradigms. Mean accuracies of only $60.4\%$, $70.0\%$, and $67.4\%$ for MI EEG classification tasks were reported in \cite{guger2003many, ahn2013high, cho2017eeg}, respectively.
This can be partially attributed to the inability of the target users in operating the MI-related BCI system, also known as "BCI-illiteracy" \cite{guger2003many, ahn2013high, vidaurre2010towards, nicolas2012brain}. The recent study in \cite{lee2019eeg} reported an average BCI-illiteracy of 53.7\% for the MI BCI paradigm, but 11.1\% and 10.2\% for ERP and SSVEP, respectively. BCI illiteracy can increase the number of outliers considerably, resulting in unreliable data.

On the other hand, SI BCI studies have reported variable classification accuracies. In \cite{reuderink2011subject, dong2020classification}, maximum mean accuracies of $67\%$ and $72\%$ were reported, while in \cite{lotte2009comparison, hosseini2019auto} a comparison of the SS and SI BCI approaches reported an accuracy of $73\%$ and $64\%$ for the former, whereas the latter achieved $77\%$ and $68\%$ accuracy.
The SI accuracy in \cite{hosseini2019auto} was increased to $84\%$ by considering various modifications, such as trial removal and partial inclusion of new user data in the training procedure. 

Low classification accuracy can result in an inappropriate control signal and consequently make the MI EEG-based BCI system fail in either the SS or SI paradigms.
BCI-illiteracy, lack of required experimentation, high between-subject variability, and small sample sizes can significantly contribute to low classification accuracy. Small sample size require the use of simple/linear statistical machine learning (ML) methods with few degrees of freedom \cite{ghane2020learning}. However, even with simple  ML methods, small sample size can pose issues due to the curse of dimensionality phenomenon \cite{verleysen2005curse, trunk1979problem} and limitations in test-train split of the data, among many other issues \cite{golugula2011evaluating, jain1997feature}. 
In addition, well-known methods like cross-validation may become inadmissible due to their high variance in MI EEG data-poor environments \cite{kohavi1995study, varoquaux2018cross, braga2004cross}.

In this paper, we performed an empirical study of the classification accuracy of MI EEG-based SS and SI BCI designs, using Linear Discriminant Analysis (LDA), Support Vector Machines (SVM), Classification and regression tree (CART), and k-Nearest Neighbors (KNN), while investigating the effect of varying training sample size on classification accuracy.
The rest of this paper is organized as follows. The basic classification methods and performance evaluation are described in Section \ref{sec:methods}. Section \ref{sec:results} contains the experimental setups and results, which are  discussed in Section~\ref{sec:discussion}. Finally, Section~\ref{sec:conclusions} contains conclusions. 


\section{METHODS} \label{sec:methods}

\subsection{Sample Data}

We used the publicly available MI EEG data provided by the GigaScience Repository \cite{lee2019eeg}. 
20 subjects (ages: 24-32 years) were seated in front of an LCD display, in a comfortable position, wearing a 62-channel EEG headset. 
Brain activity was recorded with a sampling rate of 1,000 Hz. The individuals were instructed to complete a in a single session a total of 40 trials: 20 left hand and 20 right hand imagery tasks. 

\begin{figure}[h!]
    \centering
  \includegraphics[scale=0.5]{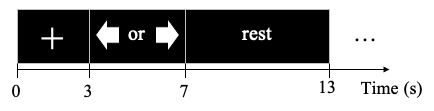}
    \caption{Experimental design for a motor imagery trial.}
    \label{fig:data_protocol}
\end{figure}

Each trial started by displaying a "+" symbol in the center of the screen which signaled the subjects to relax their muscles and prepare for the MI task. 
After 3 seconds, a right or left arrow was displayed and subjects performed the imagery task of grasping with the appropriate hand for 4 seconds. Subsequently, the screen turned black and subjects rested for 6 seconds. 
Trials were coded as a "left" or "right" class according to the displayed arrow. Figure \ref{fig:data_protocol} displays the study protocol for one trial. 

\subsection{Pre-Processing} \label{subsec:preprocess}
We restored the EEG signals from only the 20 channels that were placed on or close to the motor cortex \cite{lee2019eeg}.
Subsequently, we extracted the MI-specific segment of the signals, excluding a transitioning time between tasks. 
Then a Butterworth band-pass filter was applied to the signals to remove high-frequency noise and low-frequency artifacts and retain the brainwaves of interest \cite{ghane2015robust}, which include theta (deep relaxation and meditation state), alpha (relaxed, calm, and no-thinking state), and beta (awake and normal alert consciousness state).

\subsection{Feature Extraction and Selection}

Studies have demonstrated that features based on the power spectral density (PSD) of MI EEG signals lead to consistent and robust pattern identification \cite{oikonomou2017comparison, herman2008comparative, brodu2011comparative}.
Let $x(t)$ be the EEG signal of a recording channel. The PSD of $x(t)$, which is defined in terms of the  autocorrelation function~$R_{XX}(\tau)$, is given by: 
\begin{equation}
    S_{xx}(\omega) = \int_{-\infty}^{+\infty} R_{xx}(\tau) e^{-i\omega \tau} d\tau\,.
\label{eq:psd}
\end{equation}
The PSD is used to compute the signal power over given frequency ranges. 

Since the autocorrelation function $R_{XX}(\tau)$ of the signal $x(t)$ is not known, we estimate the PSD per recording channel using the {\em periodogram},  
\begin{equation}
\hat{S}_{xx}(\omega) = \frac{\Delta t}{T} \left| \sum_{n=0}^{N-1} x[n] e^{-i\omega n \Delta t}  \right|^2
\label{eq:perio-psd}
\end{equation} 
where $x[n] = x(n\Delta t)$ is the discrete-time version of $x(t)$ using the sampling interval $\Delta t$.

To reduce the dimensionality, we selected the maximum of the periodogram over small intervals of 10 samples. 
Next, we appended the resulting values of the channels to create a feature vector. Finally, we performed a t-test to select the most discriminating features. 
 

\subsection{Classification}

Linear Discriminant Analysis (LDA) and Support Vector Machines (SVM) have been among the most popular parametric classification algorithms for EEG-based BCI system design \cite{lee2019eeg, lotte2018review, ghane2015silent, sarmiento2014brain}. In particular, \cite{lotte2007review} highlighted that SVM has often outperformed other classifiers in the previous literature. In this section, we introduce some notation and briefly describe the LDA and SVM classification algorithms, as well as non-parametric classification algorithms, CART and k-NN.

A classifier assigns a label $y = \psi(\bx)$ to each feature vector $\bx \in R^d$, where $y=0$ and $y=1$ indicate the "right" and "left" hand imagery tasks, respectively. 
The classifier is designed using training data by $S_n=\{\bX_1, Y_1),\ldots, (\bX_n,Y_n)\}$, where $n$ is the sample size. Furthermore, $n_0$ and $n_1$ denote the sample sizes for class $y=0$ and $y=1$, respectively, with $n_0 + n_1 = n$.   

\subsubsection{Linear Discriminant Analysis (LDA)}
Assume that $p(\bx\mid y\!=\!0) = N(\bmu_0, \Sigma_0)$ and $p(\bx\mid y\!=\!1)= N(\bmu_1, \Sigma_1)$ are the class conditional densities for class $0$ and $1$, respectively. 
The unknown class means and covariance matrices are usually approximated by their maximum-likelihood estimators $\hat{\bmu}_i= \frac{1}{n_i} \sum_{j=1}^{n_i} \bX_j I_{Y_j=i}$ and $\hat{\Sigma}_i = \frac{1}{n_i - 1} \sum_{j=1}^{n_i} (\bX_j - \hat{\bmu}_i) (\bx_j - \hat{\bmu}_i)^T I_{Y_j=i}$, $i = 0,1$, respectively.
LDA makes the additional assumption that $\Sigma_0 = \Sigma_1 = \Sigma$, where the common covariance matrix $\Sigma$ can be estimated as
\begin{equation}
    \hat{\Sigma}\,=\, \frac{1}{n_0 + n_1 - 2} (n_0-1)\hat{\Sigma}_0 + (n_1 - 1)\hat{\Sigma}_0\,.
\end{equation}
The LDA classifier corresponds to plugging in the aforementioned parameter esimtators in the expression for the optimal classifier under the Gaussian distributional assumption, which yields  
\begin{equation}
\psi_n(\bx)\,=\, \left\{ 
            \begin{array}{ll}
                1, & \bw^T \bx +b > 0, \\
                0, & \textrm{otherwise},
            \end{array} 
            \right.
\end{equation}
where
\begin{equation}
    \begin{array}{l}
        \bw \,=\, \hat{\Sigma}^{-1} (\hat{\bmu_1} - \hat{\bmu_0})\,, \\
         b \,=\, \frac{1}{2} (\hat{\bmu_1} - \hat{\bmu_0})^T \hat{\Sigma}^{-1} (\hat{\bmu_0} + \hat{\bmu_1})\,.
    \end{array}
\end{equation}

\subsubsection{Support Vector Machine (SVM)}
Linear SVM is a linear discriminant with {\em maximum margins}, i.e., maximum separation between the decision boundary and the training data. It can be shown that the margin is given by $\frac{1}{||\bw||}$ and one needs to maximize that. 
The margin constraints can be written, Without loss of generality,
as $\bw \bX_j + b \geq 1$ or $\bw \bX_j + b \leq 1$, depending on whether $Y_j = 1$ or $Y_j = -1$, respectively, for $j=1,\ldots,n$ (class ``0'' here is coded by $-1$). Hence, one obtains a convex optimization problem
\begin{equation}
\begin{array}{l}
    \mbox{minimize} \ \ \ \frac{1}{2} ||\bw||^2 \\[1ex]
    \mbox{subject to } \ \ Y_i(\bw^T\bX_i +b) > 0,\ j=1,\ldots,n.
    \end{array}
\end{equation}
the solution of which is given by
\begin{equation}
    \begin{array}{l}
        \bw^* \,=\, \sum_{j\in \mathcal{S}} \lambda_j^* Y_j\bX_j\,, \\[1ex]
        b^* \,=\, -\frac{1}{n}
        \sum_{j \in \mathcal{S}} \sum_{k \in \mathcal{S}} \lambda_j^* Y_j \bX_j^T \bX_k + \frac{1}{n}\sum_{j \in \mathcal{S}} Y_j
    \end{array}
\end{equation}
where $\lambda_j^*$ are the optimal values of the Lagrange multipliers for the preceding optimization problem, for $j=1,\ldots,n$, and $\mathcal{S}$ is the set of indices of the {\em support vectors}, i.e., training points $(\bX_j,Y_j)$ for which $\lambda_j>0$. The LSVM classifier is then given by:
\begin{equation}
\psi_n(\bx)\,=\, \left\{ 
            \begin{array}{ll}
                1, & (\bw^*)^T \bx +b^* > 0, \\
                -1, & \textrm{otherwise},
            \end{array} 
            \right.
\end{equation}
In practice, slack variables are introduced into the optimization problem to allow points to violate the margin, allowing the solution of nonlinearly separable problems and avoiding overfitting.  

A non-linear SVM replaces all dot-products $\bx^T \bx'$ in the linear SVM formulation by a kernel function $k(\bx,\bx') = \Phi(\bx)^T\Phi(\bx')$. This correponds to transforming the data to a higher-dimensional space using $\Phi$ and applying a linear SVM to the transformed data. 
Here we use the radial basis function (RBF) kernel,
\begin{equation}
   k(\bx, \bx') \,=\, \exp \left(    \frac{||\bx- \bx'||^2}{2\sigma^2}     \right)
   \label{eq:rbf}
\end{equation}
where $\sigma$ is the kernel bandwidth parameter and is estimated from the observed data.

\subsubsection{Classification and regression tree (CART)}
CART is a non-parametric decision tree algorithm. At each node of the tree, CART sets a threshold on a selected feature to split the data into two groups. The threshold and feature are selected at each node such that the impurity of the children nodes is minimized. At a terminal node, or leaf node, a label is assigned according to majority vote among the training points in the leaf node. To avoid overfitting, node splitting is usually terminated early.

\subsubsection{K nearest neighbors (kNN)}
kNN is another non-parametric classification algorithm. At any point in the feature space, kNN assigns the majority label among the k nearest training points. Smoothness of the kNN classifier decision boundary increases as k increases. Too small $k$ (say $k=1$) introduces overfitting, but too large $k$ leads to underfitting. To avoid ties, odd $k$ is recommended in binary classification problems.

\section{EXPERIMENTAL RESULTS} \label{sec:results}

We report in this section the results of a comprehensive empirical comparison of the previous classification algorithms on the MI EEG data described earlier. 

Similarly to \cite{lee2019eeg}, to account for the transition time, we removed the first 1s and last 0.5s of the MI segments of the EEG signals and performed the rest of the analysis on the middle 2.5 seconds. The signals were filtered using a Butterworth band-pass filter with lower and upper cut-off frequencies of 3 and 35 Hz, respectively. 
The p-value threshold for the t-test was set to $0.05$ for the SS model, which resulted in various numbers of selected features per subject. This threshold was set on $0.005$ for the SI experiments, which resulted in total of 29 features.  

We performed several classification experiments using the LDA, RB-SVM, CART, and kNN classification algorithms. We set the minimum leaf size to 3 for CART and $k=3$ for kNN.  
Classifier generalization ability was evaluated by randomly splitting the available data into 50\% training data $S_N$ and 50\% testing data $S_M$, and using a test-set error estimator, which is the relative accuracy of the trained classifier on the testing set. 
In our SI design, the total sample size was $800$, so that the testing sample size was  $400$, which is enough to guarantee excellent test-set error estimation accuracy \cite{BragDoug:15}. 



For the training step, $n\leq 400$ sample points were randomly drawn from $S_N$. For each value of $n$, we trained the classifiers on the selected data, calculated the test set error on $S_M$, and finally computed the percent prediction accuracies as $100(1-\mbox{test set error}) \%$. The classification performances was examined with $n \in \{10, 15, 20\}$ for the SS and $n \in \{ 100, 150, 200, 250, 300, 350, 400 \}$ for the SI models.

\begin{figure*}
    \centering
    \includegraphics[scale=0.4]{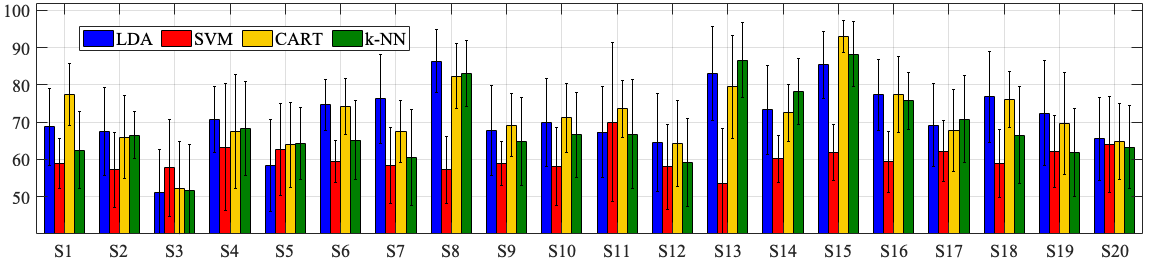}
    \caption{Barplots of the mean accuracy in SS models with $n=10$. The error bars display the standard deviations.}
    \label{fig:ss_n10}
\end{figure*}

\begin{figure*}
    \centering
    \includegraphics[scale=0.4]{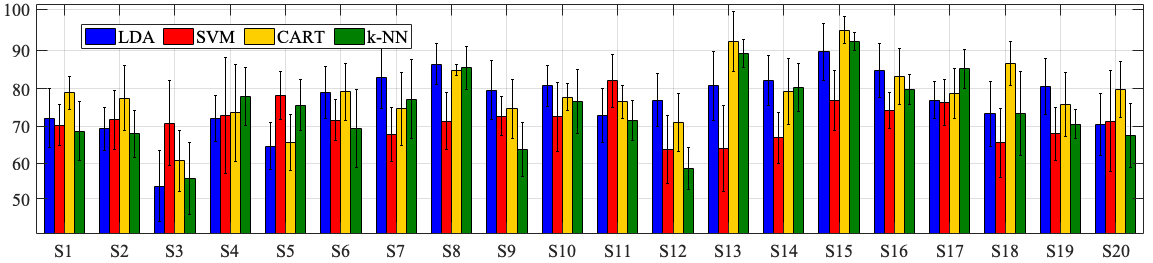}
    \caption{Barplots of the mean accuracy in SS models with $n=20$. The error bars display standard deviations.}
    \label{fig:ss_n20}
\end{figure*}

\begin{figure}
    \centering
    \includegraphics[scale= 0.45]{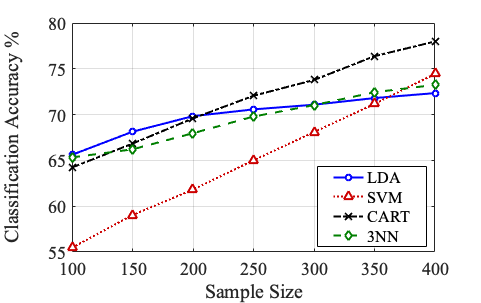}
    \caption{Comparison of the mean accuracy for SI models.}
    \label{fig:si}
\end{figure}

\begin{figure}
\vspace{-0.1 in}
\begin{subfigure} { \textwidth}
     \begin{tabular}{cc}
         Subject-independent &  \hspace{-0.3in} Subject-specific \\
            \includegraphics[width= 0.35 \textwidth, height= 0.22 \textwidth]{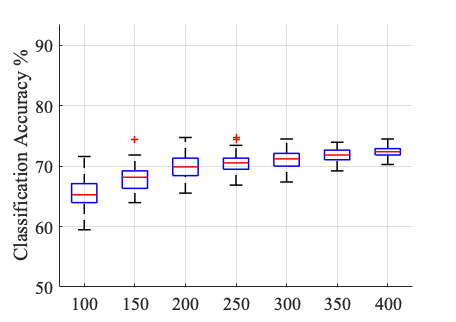} & 
            \hspace{-0.3in}
        \includegraphics[width= 0.13 \textwidth, height= 0.22 \textwidth]{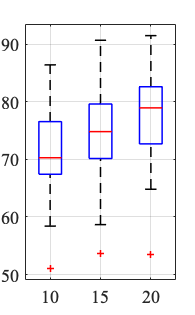}
        \end{tabular}
    \label{fig:lda_box}
\end{subfigure}

\begin{subfigure} {\textwidth}
     \begin{tabular}{cc}
            \includegraphics[width= 0.35 \textwidth, height= 0.22 \textwidth]{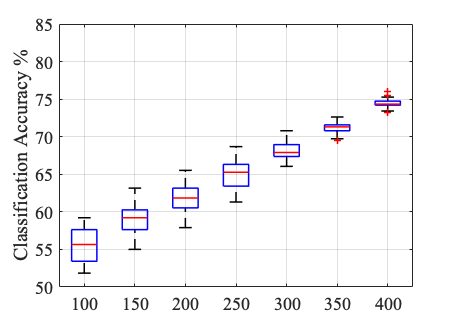} & \hspace{-0.3in}
        \includegraphics[width= 0.13 \textwidth, height= 0.22 \textwidth]{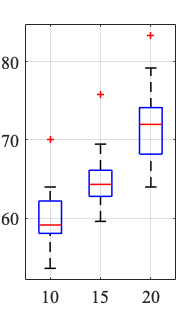}
        \end{tabular}
    \label{fig:svm_box}
\end{subfigure}

\begin{subfigure} {\textwidth}
     \begin{tabular}{cc}
            \includegraphics[width= 0.35 \textwidth, height= 0.22 \textwidth]{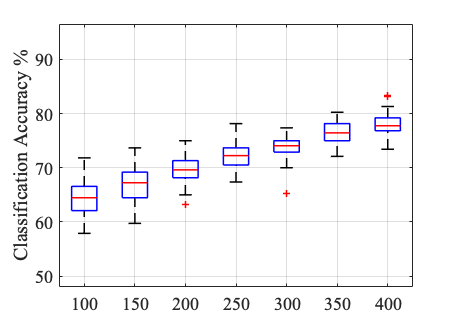} & \hspace{-0.3in}
        \includegraphics[width= 0.13 \textwidth, height= 0.22 \textwidth]{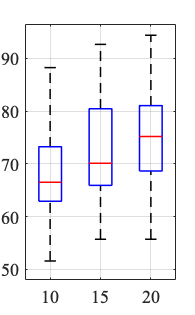}
        \end{tabular}
    \label{fig:cart_box}
\end{subfigure}

\begin{subfigure}{\textwidth}
     \begin{tabular}{cc}
            \includegraphics[width= 0.35 \textwidth, height= 0.22 \textwidth]{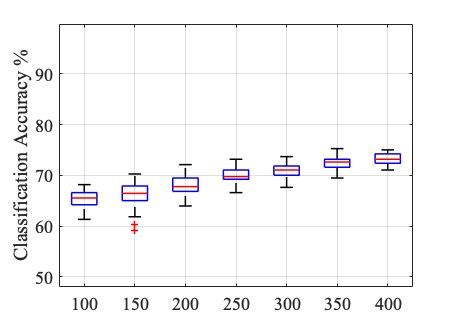} & \hspace{-0.3in}
        \includegraphics[width= 0.13 \textwidth, height= 0.22 \textwidth]{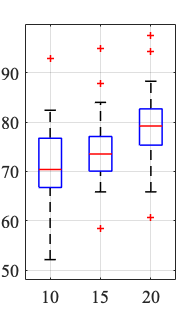}
        \end{tabular}
    \label{fig:3nn_box}
    \end{subfigure}
    \caption{Box plots for prediction accuracies of SI (left) and SS (right) designs, with classification algorithms: LDA (1\textsuperscript{st} row), SVM (2\textsuperscript{nd} row), CART (3\textsuperscript{rd} row), and 3-NN (4\textsuperscript{th} row). Horizontal axis shows sample size.}
    \label{fig:box_plots}
\end{figure}

Randomly drawn training samples introduces a randomness factor to the classifiers' training and prediction performance. To take into account this randomness factor, we repeated each experimental case 100 times.
The SS models' performance are presented in Table \ref{tab:ss} and Figures \ref{fig:ss_n10} and \ref{fig:ss_n20}.  
Table \ref{tab:si} shows the mean (std) percent accuracy of the SI models, whereas figure \ref{fig:si} compares the mean accuracy of the four classifiers for varying training sample sizes, ranging from $n=100$ to $n=400$. 
Figure \ref{fig:box_plots} displays box plots of classifier accuracy for SI and SS approaches and compares the mean accuracy and standard deviation for varying training  sample size.
\begin{table*}[]
    \centering
    \begin{tabular}{|l|ccc|ccc|ccc|ccc|ccc|}
    \hline
          Subject  & \multicolumn{3}{|c|}{LDA} & \multicolumn{3}{|c|}{SVM} & \multicolumn{3}{|c|}{CART} & \multicolumn{3}{|c|}{k-NN} & \multicolumn{3}{|c|}{$\Psi$ with highest mean accuracy}\\
     & n=10 & n=15 & n=20 & n=10 & n=15 & n=20 & n=10 & n=15 & n=20 & n=10 & n=15 & n=20 & n=10 & n=15 & n=20\\
    \hline
    S1 &    69 & 72 & 73 & 59 & 63 & 71 & 77 & 77 & 80 & 63 & 61 & 69 
        &   CART & CART & CART\\
    S2 &    67 & 70 & 70 & 56 & 67 & 72 & 66 & 71 & 78 & 67 & 68 & 68 
        &   LDA & CART & CART\\
    S3 &    51 & 54 & 53 & 57 & 62 & 71 & 52 & 57 & 61 & 52 & 56 & 56 
        &   SVM & SVM & SVM\\
    S4 &    71 & 71 & 73 & 63 & 60 & 73 & 67 & 68 & 74 & 68 & 72 & 79 
        &   LDA & k-NN & k-NN\\
    S5 &    57 & 59 & 65 & 64 & 69 & 79 & 64 & 66 & 66 & 64 & 67 & 76 
        &   SVM & SVM & SVM\\
    S6 &    75 & 78 & 80 & 59 & 64 & 72 & 74 & 75 & 80 & 65 & 70 & 70 
        &   LDA & LDA & LDA\\
    S7 &    76 & 80 & 84 & 57 & 61 & 68 & 68 & 71 & 75 & 61 & 65 & 78 
        &   LDA & LDA & LDA\\
    S8 &    86 & 91 & 88 & 56 & 63 & 72 & 82 & 84 & 86 & 83 & 86 & 87 
        &   LDA & LDA & LDA\\
    S9 &    68 & 77 & 81 & 59 & 65 & 73 & 69 & 72 & 75 & 65 & 67 & 64 
        &   CART & LDA & LDA\\
    S10 &    70 & 74 & 82 & 57 & 64 & 73 & 71 & 73 & 79 & 67 & 72 & 77 
        &   LDA & LDA & LDA\\
    S11 &    67 & 70 & 73 & 70 & 76 & 83 & 74 & 76 & 77 & 67 & 72 & 72 
        &   CART & CART & SVM\\
    S12 &    65 & 71 & 78 & 57 & 65 & 64 & 64 & 67 & 72 & 59 & 59 & 57 
        &   LDA & LDA & LDA\\
    S13 &    83 & 79 & 82 & 54 & 64 & 64 & 79 & 88 & 94 & 87 & 89 & 91 
        &   k-NN & k-NN & CART\\
    S14 &    73 & 80 & 83 & 60 & 62 & 67 & 73 & 74 & 80 & 78 & 80 & 81 
        &   k-NN & LDA & LDA\\
    S15 &    85 & 86 & 91 & 62 & 69 & 78 & 93 & 95 & 98 & 88 & 93 & 94 
        &   CART & CART & CART\\
    S16 &    77 & 80 & 86 & 59 & 65 & 75 & 77 & 77 & 85 & 76 & 81 & 81 
        &   LDA & k-NN & LDA\\
    S17 &    69 & 70 & 78 & 62 & 66 & 77 & 68 & 74 & 80 & 71 & 81 & 87 
        &   k-NN & k-NN & k-NN\\
    S18 &    77 & 75 & 74 & 59 & 60 & 66 & 76 & 78 & 88 & 67 & 70 & 74 
        &   LDA & CART & CART\\
    S19 &    72 & 76 & 82 & 62 & 66 & 68 & 70 & 72 & 77 & 62 & 69 & 71 
        &   LDA & LDA & LDA\\
    S20 &    66 & 66 & 71 & 64 & 65 & 72 & 65 & 69 & 81 & 63 & 63 & 68 
        &   LDA & CART & CART\\
    \hline
    {\bf mean} &  71   & 74  &  77  &  60 &   65  &  72  &  71  &  74  &  79  &  69  &  72   & 75 
        &    &  & \\
    {\bf std} &    9   &  8   &  9  &   4  &   4   &  5  &   8   &  8   &  8   &  9   & 10 &   10
        &    &  & \\
    \hline
    \end{tabular}
    \caption{Mean accuracy in percentage per subject for SS models with three sample sizes $n$. Numbers are rounded to the closest integer. Last three columns show the classification algorithm with the highest mean accuracy per subject per sample size.}
    \label{tab:ss}
\end{table*}
\begin{figure*}
    \centering
    \includegraphics[scale=0.4]{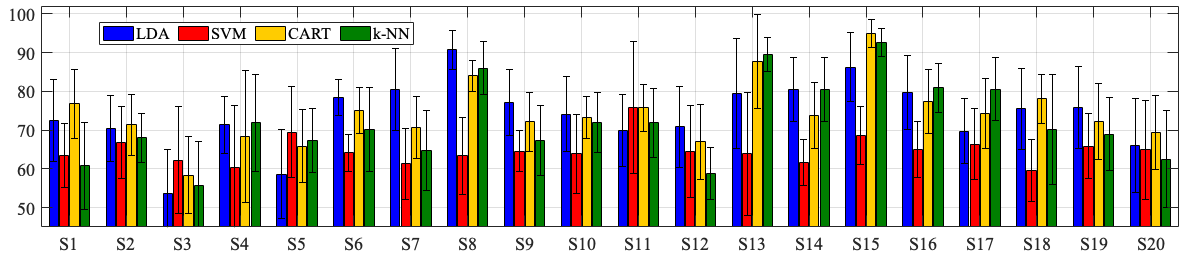}
    \caption{Barplots of the mean accuracy in SS models with $n=15$. The error bars display the standard deviations.}
    \label{fig:ss_n15}
\end{figure*}

\begin{table}[h]
    \centering
    \begin{tabular}{|c|c|c|c|c|c|}
    \hline
        $n$ & LDA & SVM & CART & k-NN & Max(acc) \\
        \hline
        100 &    66 (3) & 55 (2) & 65 (2) & 66 (3) & LDA \\
        150 &    69 (2) & 59 (1) & 68 (3) & 66 (2) & LDA \\
        200 &    69 (2) & 62 (1) & 71 (3) & 69 (2) & CART \\
        250 &    71 (2) & 65 (1) & 72 (3) & 70 (2) & CART \\
        300 &    72 (2) & 68 (1) & 74 (2) & 72 (2) & CART \\
        350 &    72 (1) & 71 (1) & 77 (3) & 73 (2) & CART \\
        400 &    73 (1) & 74 (1) & 78 (2) & 74 (1) & CART \\
        \hline
    \end{tabular}
    \caption{The mean (std) accuracies  in percentage for subject independent models. The last column shows the classification algorithm with the highest test-set prediction accuracy.}
    \label{tab:si}
\end{table}

\section{DISCUSSION} \label{sec:discussion}
Our experiments lead to several interesting conclusions.
For SI design, as displayed in figure \ref{fig:si}, in the small sample-size region ($100 \leq n \leq 200$), LDA was the best performing classifier. CART beats all other classifiers if $200 \leq n \leq 400$ samples are provided. SVM was the most robust (lowest std) in both SS and SI paradigms. The steep improvement in SVM accuracy leads to the hypothesis that it may achieve the highest accuracy if the training sample size is large enough.

For the SS design, as illustrated by Figures \ref{fig:ss_n10} - \ref{fig:ss_n20} and Table \ref{tab:ss}, increasing sample size improves the classification performance in terms of mean prediction accuracy, although it may be impracticable to collect many EEG recordings per subject.  While SVM was the best performing classifier for only a few subjects (S3, S5, and S11 when n=20), the other three classifiers compete for the highest SS accuracy. This behaviour suggests a hypothesis that a hybrid model may be beneficial for SS BCI systems, where a specific classification algorithm is selected according to the maximum accuracy of the test set for that single individual.   

The side-by-side box plots in figures \ref{fig:box_plots} show that, for all classification algorithms, in contrast to the SI models, the variance of the trained SS classifiers is not reduced with an increment in training sample size. Indeed, the variance is increased noticeably in CART and SVM. This variability is expected at least due to the high between-subject differences in EEG signals, and suggests that a trained BCI may not be appropriate for a totally new user. It is worthy to note that the study in \cite{hosseini2019auto} also demonstrated improved accuracy by including a few samples from new users in training a SI BCI.

Overall, the results indicate that SS designs, and possibly a hybrid SS model, may be more appropriate for personalized MI BCI if collecting enough samples from a specific subject is feasible. 
In case a recording session is inconvenient for a subject, particularly for mentally or physically challenged people, a SI BCI may be used. Such SI BCI should be trained on a sufficiently large training data set that is collected from a large group of subjects covering between-subject variability.
Comparing the box plots, it can be seen that classification mean accuracy in SI BCI even with a  relatively large sample size ($n=400$) is close to that of SS BCI with $n=20$. However, the low variance of SI BCI acccuracy may make it more applicable for a multi-user system.  

\section{CONCLUSIONS}\label{sec:conclusions}

This paper presents our evaluation of the subject-specific (SS) and subject-independent (SI) paradigms using two popular classification algorithms in MI BCI design, LDA and SVM, in addition to the non-parametric algorithms k-NN and CART. 
Our results show that for SI BCI, LDA beats other classification algorithms in the presence of small sample size. CART outperforms the rest for a relatively large sample size, whereas we hypothesize that SVM achieves the highest accuracy with a sufficiently large sample size. It is important to stress that generalization performance should be evaluated on a relatively large and independent test set, which was done here. 

For SS models, although LDA and CART performed well for majority of subjects, the best classifier performance depends on the subject, which suggests the applicability of a hybrid model. 
As was demonstrated in\cite{cantillo2014approach} for gender-specific BCIs, considering the subject gender may improve the performance of SI designs. We leave the investigation of gender-specific SI BCI performance to future work.
Another interesting research problem is to investigate the EEG recording channels with the least discriminating information, for example by using a sequential backward search. 
In addition, deep learning methods, particularly convolutional neural networks (CNNs), are able to process brain maps as input images, but may need larger sample size for training a network with good generalization abilities. A future experiment may study the performance of CNNs for SI BCI considering varying sample sizes.   

\bibliographystyle{IEEEtran.bst} 
\bibliography{IEEEabrv.bib, myReferences.bib}

\begin{thebibliography}{10}
\providecommand{\url}[1]{#1}
\csname url@samestyle\endcsname
\providecommand{\newblock}{\relax}
\providecommand{\bibinfo}[2]{#2}
\providecommand{\BIBentrySTDinterwordspacing}{\spaceskip=0pt\relax}
\providecommand{\BIBentryALTinterwordstretchfactor}{4}
\providecommand{\BIBentryALTinterwordspacing}{\spaceskip=\fontdimen2\font plus
\BIBentryALTinterwordstretchfactor\fontdimen3\font minus
  \fontdimen4\font\relax}
\providecommand{\BIBforeignlanguage}[2]{{%
\expandafter\ifx\csname l@#1\endcsname\relax
\typeout{** WARNING: IEEEtran.bst: No hyphenation pattern has been}%
\typeout{** loaded for the language `#1'. Using the pattern for}%
\typeout{** the default language instead.}%
\else
\language=\csname l@#1\endcsname
\fi
#2}}
\providecommand{\BIBdecl}{\relax}
\BIBdecl

\bibitem{pfurtscheller2001motor}
G.~Pfurtscheller and C.~Neuper, ``Motor imagery and direct brain-computer
  communication,'' \emph{Proceedings of the IEEE}, vol.~89, no.~7, pp.
  1123--1134, 2001.

\bibitem{jeannerod1995mental}
M.~Jeannerod, ``Mental imagery in the motor context,'' \emph{Neuropsychologia},
  vol.~33, no.~11, pp. 1419--1432, 1995.

\bibitem{decety1994mapping}
J.~Decety, D.~Perani, M.~Jeannerod, V.~Bettinardi, B.~Tadary, R.~Woods, J.~C.
  Mazziotta, and F.~Fazio, ``Mapping motor representations with positron
  emission tomography,'' \emph{Nature}, vol. 371, no. 6498, pp. 600--602, 1994.

\bibitem{kevric2017comparison}
J.~Kevric and A.~Subasi, ``Comparison of signal decomposition methods in
  classification of eeg signals for motor-imagery bci system,''
  \emph{Biomedical Signal Processing and Control}, vol.~31, pp. 398--406, 2017.

\bibitem{guger2009many}
C.~Guger, S.~Daban, E.~Sellers, C.~Holzner, G.~Krausz, R.~Carabalona,
  F.~Gramatica, and G.~Edlinger, ``How many people are able to control a
  p300-based brain--computer interface (bci)?'' \emph{Neuroscience letters},
  vol. 462, no.~1, pp. 94--98, 2009.

\bibitem{lee2019eeg}
M.-H. Lee, O.-Y. Kwon, Y.-J. Kim, H.-K. Kim, Y.-E. Lee, J.~Williamson,
  S.~Fazli, and S.-W. Lee, ``Eeg dataset and openbmi toolbox for three bci
  paradigms: an investigation into bci illiteracy,'' \emph{GigaScience},
  vol.~8, no.~5, p. giz002, 2019.

\bibitem{guger2003many}
C.~Guger, G.~Edlinger, W.~Harkam, I.~Niedermayer, and G.~Pfurtscheller, ``How
  many people are able to operate an eeg-based brain-computer interface
  (bci)?'' \emph{IEEE transactions on neural systems and rehabilitation
  engineering}, vol.~11, no.~2, pp. 145--147, 2003.

\bibitem{ahn2013high}
M.~Ahn, H.~Cho, S.~Ahn, and S.~C. Jun, ``High theta and low alpha powers may be
  indicative of bci-illiteracy in motor imagery,'' \emph{PloS one}, vol.~8,
  no.~11, p. e80886, 2013.

\bibitem{cho2017eeg}
H.~Cho, M.~Ahn, S.~Ahn, M.~Kwon, and S.~C. Jun, ``Eeg datasets for motor
  imagery brain--computer interface,'' \emph{GigaScience}, vol.~6, no.~7, p.
  gix034, 2017.

\bibitem{vidaurre2010towards}
C.~Vidaurre and B.~Blankertz, ``Towards a cure for bci illiteracy,''
  \emph{Brain topography}, vol.~23, no.~2, pp. 194--198, 2010.

\bibitem{nicolas2012brain}
L.~F. Nicolas-Alonso and J.~Gomez-Gil, ``Brain computer interfaces, a review,''
  \emph{sensors}, vol.~12, no.~2, pp. 1211--1279, 2012.

\bibitem{reuderink2011subject}
B.~Reuderink, J.~Farquhar, M.~Poel, and A.~Nijholt, ``A subject-independent
  brain-computer interface based on smoothed, second-order baselining,'' in
  \emph{2011 Annual International Conference of the IEEE Engineering in
  Medicine and Biology Society}.\hskip 1em plus 0.5em minus 0.4em\relax IEEE,
  2011, pp. 4600--4604.

\bibitem{dong2020classification}
E.~Dong, C.~Yue, and S.~Du, ``Classification of subject-independent motor
  imagery eeg based on relevance vector machine,'' in \emph{2020 IEEE
  International Conference on Mechatronics and Automation (ICMA)}.\hskip 1em
  plus 0.5em minus 0.4em\relax IEEE, 2020, pp. 67--71.

\bibitem{lotte2009comparison}
F.~Lotte, C.~Guan, and K.~K. Ang, ``Comparison of designs towards a
  subject-independent brain-computer interface based on motor imagery,'' in
  \emph{2009 Annual International Conference of the IEEE Engineering in
  Medicine and Biology Society}.\hskip 1em plus 0.5em minus 0.4em\relax IEEE,
  2009, pp. 4543--4546.

\bibitem{hosseini2019auto}
S.~M. Hosseini, M.~Bavafa, and V.~Shalchyan, ``An auto-adaptive approach
  towards subject-independent motor imagery bci,'' in \emph{2019 26th National
  and 4th International Iranian Conference on Biomedical Engineering
  (ICBME)}.\hskip 1em plus 0.5em minus 0.4em\relax IEEE, 2019, pp. 167--171.

\bibitem{ghane2020learning}
P.~Ghane and G.~Hossain, ``Learning patterns in imaginary vowels for an
  intelligent brain computer interface (bci) design,'' \emph{arXiv preprint
  arXiv:2010.12066}, 2020.

\bibitem{verleysen2005curse}
M.~Verleysen and D.~Fran{\c{c}}ois, ``The curse of dimensionality in data
  mining and time series prediction,'' in \emph{International work-conference
  on artificial neural networks}.\hskip 1em plus 0.5em minus 0.4em\relax
  Springer, 2005, pp. 758--770.

\bibitem{trunk1979problem}
G.~V. Trunk, ``A problem of dimensionality: A simple example,'' \emph{IEEE
  Transactions on pattern analysis and machine intelligence}, no.~3, pp.
  306--307, 1979.

\bibitem{golugula2011evaluating}
A.~Golugula, G.~Lee, and A.~Madabhushi, ``Evaluating feature selection
  strategies for high dimensional, small sample size datasets,'' in \emph{2011
  Annual International conference of the IEEE engineering in medicine and
  biology society}.\hskip 1em plus 0.5em minus 0.4em\relax IEEE, 2011, pp.
  949--952.

\bibitem{jain1997feature}
A.~Jain and D.~Zongker, ``Feature selection: Evaluation, application, and small
  sample performance,'' \emph{IEEE transactions on pattern analysis and machine
  intelligence}, vol.~19, no.~2, pp. 153--158, 1997.

\bibitem{kohavi1995study}
R.~Kohavi \emph{et~al.}, ``A study of cross-validation and bootstrap for
  accuracy estimation and model selection,'' in \emph{Ijcai}, vol.~14,
  no.~2.\hskip 1em plus 0.5em minus 0.4em\relax Montreal, Canada, 1995, pp.
  1137--1145.

\bibitem{varoquaux2018cross}
G.~Varoquaux, ``Cross-validation failure: small sample sizes lead to large
  error bars,'' \emph{Neuroimage}, vol. 180, pp. 68--77, 2018.

\bibitem{braga2004cross}
U.~M. Braga-Neto and E.~R. Dougherty, ``Is cross-validation valid for
  small-sample microarray classification?'' \emph{Bioinformatics}, vol.~20,
  no.~3, pp. 374--380, 2004.

\bibitem{ghane2015robust}
P.~Ghane, G.~Hossain, and A.~Tovar, ``Robust understanding of eeg patterns in
  silent speech,'' in \emph{2015 National Aerospace and Electronics Conference
  (NAECON)}.\hskip 1em plus 0.5em minus 0.4em\relax IEEE, 2015, pp. 282--289.

\bibitem{oikonomou2017comparison}
V.~P. Oikonomou, K.~Georgiadis, G.~Liaros, S.~Nikolopoulos, and
  I.~Kompatsiaris, ``A comparison study on eeg signal processing techniques
  using motor imagery eeg data,'' in \emph{2017 IEEE 30th international
  symposium on computer-based medical systems (CBMS)}.\hskip 1em plus 0.5em
  minus 0.4em\relax IEEE, 2017, pp. 781--786.

\bibitem{herman2008comparative}
P.~Herman, G.~Prasad, T.~M. McGinnity, and D.~Coyle, ``Comparative analysis of
  spectral approaches to feature extraction for eeg-based motor imagery
  classification,'' \emph{IEEE Transactions on Neural Systems and
  Rehabilitation Engineering}, vol.~16, no.~4, pp. 317--326, 2008.

\bibitem{brodu2011comparative}
N.~Brodu, F.~Lotte, and A.~L{\'e}cuyer, ``Comparative study of band-power
  extraction techniques for motor imagery classification,'' in \emph{2011 IEEE
  Symposium on Computational Intelligence, Cognitive Algorithms, Mind, and
  Brain (CCMB)}.\hskip 1em plus 0.5em minus 0.4em\relax IEEE, 2011, pp. 1--6.

\bibitem{lotte2018review}
F.~Lotte, L.~Bougrain, A.~Cichocki, M.~Clerc, M.~Congedo, A.~Rakotomamonjy, and
  F.~Yger, ``A review of classification algorithms for eeg-based
  brain--computer interfaces: a 10 year update,'' \emph{Journal of neural
  engineering}, vol.~15, no.~3, p. 031005, 2018.

\bibitem{ghane2015silent}
P.~Ghane, ``Silent speech recognition in eeg-based brain computer interface,''
  Ph.D. dissertation, 2015.

\bibitem{sarmiento2014brain}
L.~Sarmiento, P.~Lorenzana, C.~Cortes, W.~Arcos, J.~Bacca, and A.~Tovar,
  ``Brain computer interface (bci) with eeg signals for automatic vowel
  recognition based on articulation mode,'' in \emph{5th ISSNIP-IEEE Biosignals
  and Biorobotics Conference (2014): Biosignals and Robotics for Better and
  Safer Living (BRC)}.\hskip 1em plus 0.5em minus 0.4em\relax IEEE, 2014, pp.
  1--4.

\bibitem{lotte2007review}
F.~Lotte, M.~Congedo, A.~L{\'e}cuyer, F.~Lamarche, and B.~Arnaldi, ``A review
  of classification algorithms for eeg-based brain--computer interfaces,''
  \emph{Journal of neural engineering}, vol.~4, no.~2, p.~R1, 2007.

\bibitem{BragDoug:15}
U.~Braga-Neto and E.~Dougherty, \emph{Error Estimation for Pattern
  Recognition}.\hskip 1em plus 0.5em minus 0.4em\relax New York: Wiley, 2015.

\bibitem{cantillo2014approach}
J.~Cantillo-Negrete, J.~Gutierrez-Martinez, R.~I. Carino-Escobar,
  P.~Carrillo-Mora, and D.~Elias-Vinas, ``An approach to improve the
  performance of subject-independent bcis-based on motor imagery allocating
  subjects by gender,'' \emph{Biomedical engineering online}, vol.~13, no.~1,
  p. 158, 2014.

\end{thebibliography}

\end{document}